\documentclass[letterpaper]{article} 
\usepackage{aaai2026}
\usepackage{booktabs} 
\usepackage{times}  
\usepackage{helvet}  
\usepackage{courier}  
\usepackage[hyphens]{url}  
\usepackage{graphicx} 
\usepackage{natbib}  
\usepackage{caption} 
\usepackage{algorithm}
\usepackage{algorithmic}
\usepackage{multirow}
\usepackage{subcaption} 
\usepackage{array}

\usepackage{newfloat}
\usepackage{listings}
\DeclareCaptionStyle{ruled}{labelfont=normalfont,labelsep=colon,strut=off} 
\floatstyle{ruled}
\newfloat{listing}{tb}{lst}{}
\floatname{listing}{Listing}
\usepackage{amsmath}
\usepackage{amsfonts}
\newtheorem{definition}{Definition}

\title{TTF: A Trapezoidal Temporal Fusion Framework for LTV Forecasting in Douyin}

\author{
    Yibing Wan\textsuperscript{\rm 2}\equalcontrib\thanks{Work done while the author was an intern at Douyin Group.},
    Zhengxiong Guan\textsuperscript{\rm 1}\equalcontrib,
    Chaoli Zhang\textsuperscript{\rm 3}\thanks{Corresponding author},
    Xiaoyang Li\textsuperscript{\rm 1},
    Lai Xu\textsuperscript{\rm 1},
    Beibei Jia\textsuperscript{\rm 1},
    Zhenzhe Zheng\textsuperscript{\rm 2},
    Fan Wu\textsuperscript{\rm 2}
}

\affiliations{
    \textsuperscript{\rm 1}Douyin Group\\
    \textsuperscript{\rm 2}Shanghai Jiao Tong University\\
    \textsuperscript{\rm 3}Zhejiang Normal University\\
    wan\_yib@sjtu.edu.cn,
    {guanzhengxiong,lixiaoyang.27,xulai,jiabeibei}@bytedance.com,
    chaolizcl@zjnu.edu.cn,
    zhengzhenzhe@sjtu.edu.cn, 
    fwu@cs.sjtu.edu.cn
}

\usepackage{bibentry}

\begin{document}

\maketitle

\begin{abstract}
In the user growth scenario, Internet companies invest heavily in paid acquisition channels to acquire new users. But sustainable growth depends on acquired users' generating lifetime value (LTV) exceeding customer acquisition cost (CAC). In order to maximize LTV/CAC ratio, it is crucial to predict channel-level LTV in an early stage for further optimization of budget allocation. The LTV forecasting problem is significantly different from traditional time series forecasting problems, and there are three main challenges. Firstly, it is an unaligned multi-time series forecasting problem that each channel has a number of LTV series of different activation dates. Secondly, to predict in the early stage, it faces the imbalanced short-input long-output (SILO) challenge. Moreover, compared with the commonly used time series datasets, the real LTV series are volatile and non-stationary, with more frequent fluctuations and higher variance. In this work, we propose a novel framework called Trapezoidal Temporal Fusion (TTF) to address the above challenges. We introduce a trapezoidal multi-time series module to deal with data unalignment and SILO challenges, and output accurate predictions with a multi-tower structure called MT-FusionNet. The framework has been deployed to the online system for Douyin. Compared to the previously deployed online model, $\mathrm{MAPE_{p}}$ decreased by 4.3\%, and $\mathrm{MAPE_{a}}$ decreased by 3.2\%, where $\mathrm{MAPE_{p}}$ denotes the point-wise MAPE of the LTV curve and $\mathrm{MAPE_{a}}$ denotes the MAPE of the aggregated LTV.

\end{abstract}


\section{Introduction}



The rapid user growth of short-form video platforms like Douyin has intensified market competition, making multi-channel paid acquisition a critical strategy for further expansion. Given finite budgets, the objective is to maximize overall profitability by optimally allocating budgets across these channels. Allocation decisions are based on LTV/CAC ratio, where CAC is the Customer Acquisition Cost and LTV is Lifetime Value. The core challenge lies in calculating LTV/CAC ratio of each channel: while the CAC per channel is planned, the LTV of each channel is uncertain at the point of acquisition. Therefore, accurate forecasting of the LTV of each channel is essential.

Recent research ~\cite{li2022billion,cowan2023modelling,xing2021learning,bauer2021improved} on LTV prediction falls into two main categories: (1) user-level prediction~\cite{wu2023contrastive, zhao2023perCLTV}, which leverages deep learning models to estimate individual LTV from user-specific features, and (2) channel-level prediction~\cite{weng2024optdist}, which treats LTV as a time series by aggregating in acquisition channels. However, in our scenario, where newly acquired users exhibit sparse behavioral data, user-level predictions are unreliable. Instead, channel-level LTV, representing the average value of users per acquisition channel, offers greater stability and predictive feasibility. Consequently, we frame LTV forecasting as a multi-time series forecasting problem, focusing on channel-level estimation to support scalable and actionable growth strategies.

There are three principal challenges: unaligned multi-time series, short-input long-output, and volatile series. Firstly, the unaligned multi-time series challenge stems from varying activation dates of series, resulting in unaligned start times and unequal lengths. This temporal unalignment complicates joint modeling of the series and the lack of temporal synchronization makes precise prediction even more difficult. Secondly, the task faces an imbalanced Short-Input Long-Output (SILO) challenge. Since it is crucial to obtain an accurate prediction of LTV in the early stage, the task demands predicting the long-term LTV up to 330 time steps from merely 30 days of observed retention data. With limited early-stage information, models struggle to capture meaningful patterns for accurate forecasting.  
Thirdly, the volatile and non-stationary series exhibit erratic and frequent fluctuations. Rapid and non-systematic variations in real LTV series produce jagged temporal patterns lacking smooth transitions, which pose obstacles to forecasting~\cite{liu2022non, hyndman2018forecasting}.

To address these challenges, we propose a framework called Trapezoidal Temporal Fusion (TTF). Our method comprises three key components. First, we propose a novel data construction method utilizing longer series information with earlier activation date, called trapezoidal multi-time series module, to solve the challenges caused by imbalanced input and output. Second, we propose a multi-tower structure called MT-FusionNet, enhancing the robustness of the model to heterogeneous data. Third, we propose a utilitarian loss compatible with the trapezoidal multi-time series module, which can improve prediction accuracy.
The model has been deployed in a production environment and integrated into an online system, where it has demonstrated better performance in practical applications.
Our contributions are as follows. 

\begin{itemize}
\item We propose a novel data structure for unaligned multi-time series called the trapezoidal multi-time series module.
\item A novel MT-FusionNet structure is proposed to enhance the robustness of the model to heterogeneous data. 
\item We introduce a utilitarian loss that is compatible with the trapezoidal multi-time series module, thereby enhancing the accuracy of the model. 
\item The framework has been deployed and is serving Douyin's online business. Both experiments and the online results have shown the effectiveness of our approach.
\end{itemize}

\section{Preliminaries}






This section formalizes the core concepts and notation for LTV forecasting in paid acquisition.
Based on the definitions, we introduce two core challenges:
(1) unaligned multi-time series, 
and (2) the short-input long-output (SILO) problem.
These challenges distinguish our task from traditional multivariate time series forecasting and motivate our proposed approach.

\begin{figure}[t]
\centering
\includegraphics[width=\linewidth]{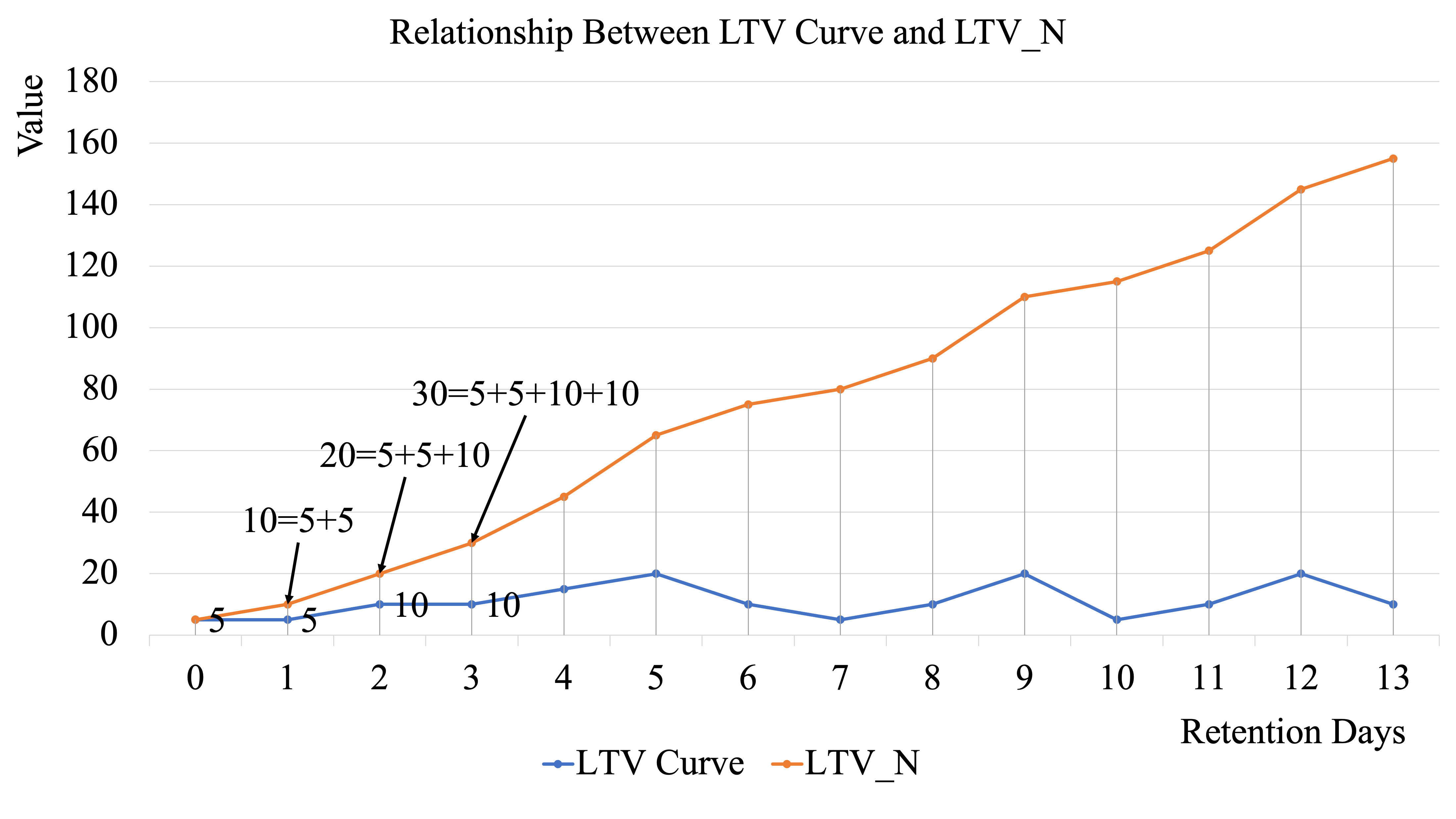}
\caption{Relationship between LTV Curve and LTV\_N. LTV Curve describe the value that generated by acquired users through a specific channel on a specific activation date, calculated on each retention day. LTV\_N stands for the cumulative sum of LTV Curve over N retention days. }
\label{fig: ltv}
\end{figure}

\begin{definition}[Activation Date]
 The activation date refers to the specific date when new users of a channel are acquired to the platform.
\end{definition}

\begin{definition}[Retention Day]
 Retention day is typically expressed as Day N retention (e.g., Day 1, Day 7, Day 30 retention), where N represents the number of days since the user's activation date. In particular, the activation date is Day 0 retention.
\end{definition}

Each user will have two identifiers: channel ID and activation date. For users activated on the same day by the same channel, they will have the exact same identifier and will be grouped together. The size of each group is defined as the number of new users activated on date $t$ on channel $c$, denoted as $U_c^t$ (Channel Activated User Count).

We track the average LTV of a group instead of individual users. LTV is generated daily from the activation date, producing a time series. We refer to this time series as the LTV curve.

\begin{definition}[LTV Curve]
The average LTV of users acquired through a specific channel on a given activation date, aggregated by retention day. We denote $LTV_c^t[i]$ as the LTV curve value of the channel $c$ and the activation date $t$ on the retention day $i^{th}$. 
\end{definition}

Based on the LTV curve, the total life value $LTV_c^t\_N$ of the group of users from the activation day to the $N^{th}$ retention day can be calculated.

\begin{definition}[LTV\_N] As Fig~\ref{fig: ltv} shows, the cumulative sum of LTV Curve over N retention days by users acquired through channel $c$ on activation date $t$:
$$ LTV_c^t\_N = \sum_{i=0}^{N} LTV_c^t[i].$$
\end{definition}

\begin{table*}[t] 
\centering
\caption{Summary of Notations}
\label{tab:notation}
\begin{tabular}{cl}
\toprule
\textbf{Symbol} & \textbf{Description} \\
\midrule
$LTV_c^t$, $LTV_c^t[i]$ & LTV curve of users activated on date $t$ in channel $c$. The $i^{th}$ value of LTV curve.\\
$LTV_c^t\_N$ & The cumulative sum of LTV value over N retention days of users activated on date $t$ in channel $c$. \\
$U_c^t$ & Channel Activated User Count. Number of new users activated on date $t$ in channel $c$. \\
$m$ & Minimum info length in a trapezoidal input window. \\
$n$ & Output length. \\
$\mathbf{M}_c^t$ & The trapezoidal input window of channel $c$ and start day $t$. \\
$s$ & Stride. The difference between the activation dates of two adjacent series in a trapezoidal input window. \\
$l$ & Input length. The length of series in a trapezoidal input window. \\
$l^{in}_j$ & The info length of $j^{th}$ series in a trapezoidal input window. \\
$k$ & Window size. Number of LTV series in a trapezoidal input window. \\
$\mathbf{M}_c^t[i,j]$ & The $i^{th}$ value of the $j^{th}$ LTV series in the trapezoidal input window of channel $c$ and start day $t$. \\
$\mathbf{Y}_c^t$, $\mathbf{\hat{Y}}_c^t$ & Real and predict output of input $\mathbf{M}_c^t$.\\
$\mathbf{w}$, $w_q$ & The moving average scales. The $q^{th}$ moving average scale. \\
$\mathbf{S}(w_q)_c^t$ & The smoothed window of moving average scale $w_q$, channel $c$ and start day $t$. \\
$d$ & Number of towers in MT-FusionNet.\\
\bottomrule
\end{tabular}
\end{table*}

Due to the distinctiveness of LTV, we face the following two challenges: 
\paragraph{Unaligned Multi-Time Series} From a channel perspective, there are many unaligned series with different lengths and activation dates. On a specific date T, a channel will have different users with activation dates of T, T-1, ..., T-k, corresponding to LTV curves of length 0, 1, ..., k. Moreover, since the activation dates of the series are different, the start times of the series are not aligned. Therefore, our scenario is significantly different from the traditional multivariate time series forecasting problem. To the best of our knowledge, this is the first work to discuss the unaligned multi-time series forecasting, and existing models cannot handle such scenarios.

\paragraph{Imbalanced Short-Input Long-Output} In the context of Douyin's user growth, it is crucial to obtain a relatively accurate prediction of LTV at an early stage. The objective is to predict the LTV of user retention up to one year based on the first 30 retention days. Specifically, with a 30-step look-back window, the model predicts the next 330 time steps. 


The task faces an imbalanced Short-Input Long-Output (SILO) challenge. Traditional time series forecasting tasks have an output comparable with or much shorter than the input, while the output is ten times longer than the input in our task. The SILO challenge brings many difficulties to existing time series forecasting models. Firstly, due to the short length of the input series, it is susceptible to abnormal fluctuations that the trend information, such as the mean and the median of series may not be accurate. 
Moreover, periodic information is crucial for time series prediction, but too short input may not encompass a complete cycle. For example, if a sequence has an annual periodicity, but the input length is only 180 days, the short input will not provide much information about the periodicity. It is challenging to generate long and stable outputs with inadequate information.


\section{Trapezoidal Multi-Time Series Module}



\begin{figure}[t]
\centering
\includegraphics[width=\linewidth]
{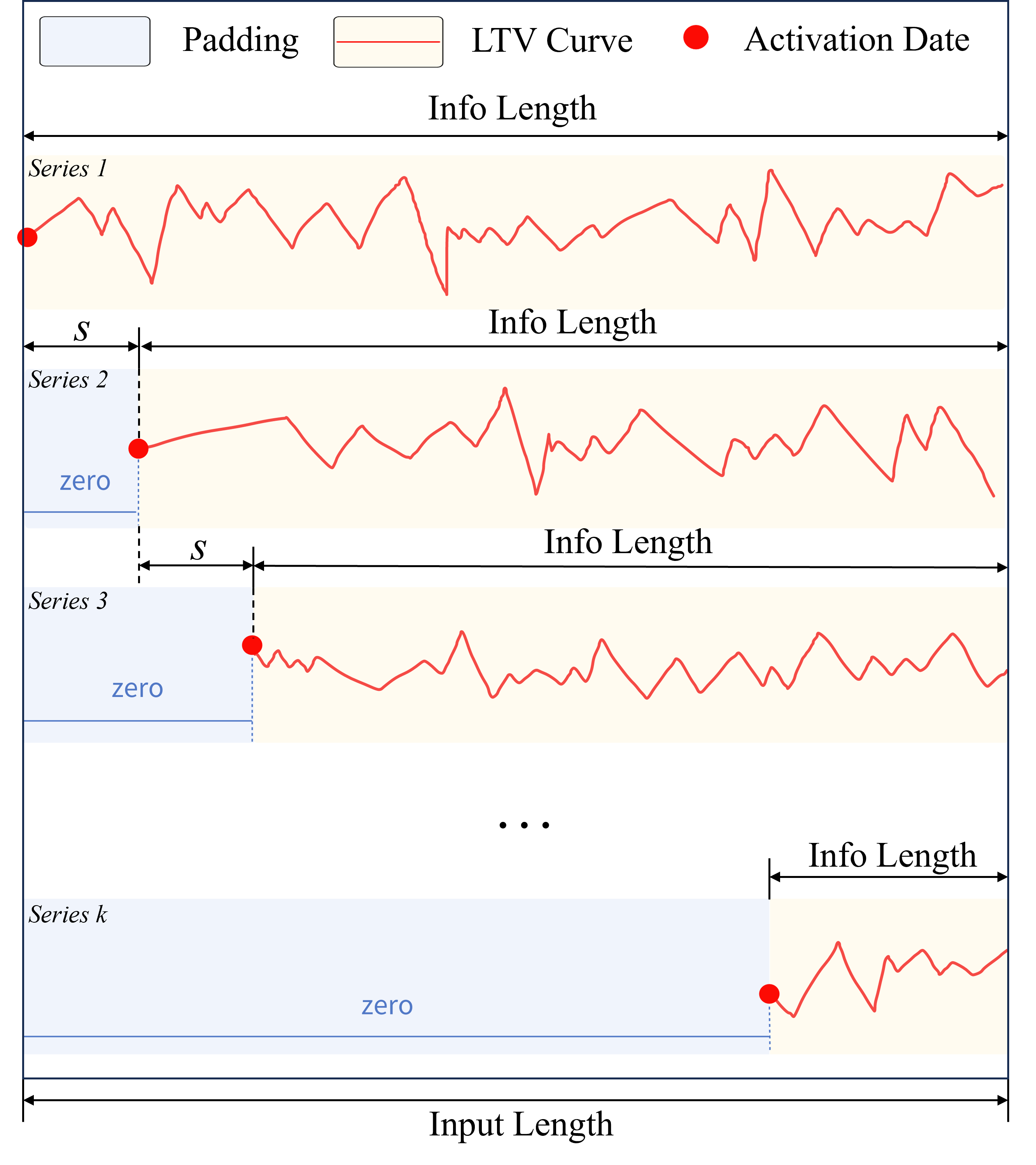} 
\caption{Trapezoidal Multi-Time Series Module. This figure shows a trapezoidal input window in a certain channel. There are $k$ series in the window with increasing activation dates and decreasing info lengths. The number of days between two adjacent activation dates are denoted as stride $s$.}
\label{fig: trapezoidal input}
\end{figure}

Formally, for a series with activation date $T$, our task is to predict the series $n$ days ahead on day $T+m$, which is to forecast an output of $n$ days with input of $m$ days. As $n$ is much larger than $m$ (e.g., $n \geq 10m$), it is a Short-Input Long-Output (SILO) time series forecasting problem. In order to solve the challenges caused by short-input long-output, we propose a novel data construction method utilizing unaligned multi-time series information, called the trapezoidal multi-time series module, as shown in figure~\ref{fig: trapezoidal input}. It leverages multi-time series information of varying lengths to supplement the insufficient input.

\subsection{Input Structure}

The trapezoidal multi-time series module concatenates $k$ sequences from the same channel in descending order of activation date into a two-dimensional matrix. The activation date of the first series is denoted as the start day. Given the length of the series $l$, the trapezoid input of channel $c$ with the start day $t$ has the shape $\mathbf{M}_c^t \in \mathbb{R}^{l \times k}$, covering actual days from day $t$ to day $t+l$. We refer to each such two-dimensional matrix as a window, and the window size is represented by the number of series $k$ in the window. The difference between the activation dates of two adjacent series is the stride $s > 0$. 

We distinguish between input length and info length. The two-dimensional window we construct is a single input, and the length of all series in this window is the same, called the input length. However, the length of valid information in a series varies. The length of the non-zero part is called info length. 

The motivation is to exploit longer information from former series. The last series $\mathbf{M}_c^t[:, k]$ with latest activation date holds the shortest info length $m$. The info length $l^{in}_j$ of series $\mathbf{M}_c^t[:, j] \ (1 \leq j\leq k)$ can be calculated by stride: 
\begin{equation}
l^{in}_j = m+s*(k-j). 
\end{equation}

We take the longest info length as the series length (input length): 
\begin{equation}
l = m+s*(k-1). 
\end{equation}

All series within a window are aligned by the actual date, and for the part before the activation date, we fill it with zeros. The $i^{th} (1 \leq i \leq l)$ value of the $j^{th} (1 \leq j \leq k)$ LTV series in the window starting on day $t$ is:

\begin{equation}
\mathbf{M}^{t}_{c}[i,j]= 
\begin{cases}
0, & 0 < i < s*(j-1), \\
LTV^{t+s*(j-1)}_c[i], & s*(j-1) \leq i \leq l,
\end{cases}
\end{equation}
where $LTV^{t+s*(j-1)}_c[i]$ is the $i^{th}$ value of LTV series in channel $c$ with activation date $t+s*(j-1)$. It actually refers to the LTV on $i^{th}$ retention day of users activating on date $t+s*(j-1)$ in the channel.

Though the info length of series in the same window varies, the output length is set to be equivalent. To predict future $n$ steps for all $k$ series in the window $\mathbf{M}_c^t$, the output is also a two-dimensional matrix $\mathbf{\hat{Y}}_c^t \in \mathbb{R}^{n \times k}$.

\subsection{Comparison with Traditional Multivariate Time Series Forecasting}
\label{sec: input}
The problem of multivariate time series forecasting has been extensively studied ~\cite{nietime, liuitransformer, li2023mts}. Multivariate models are expected to enhance prediction accuracy by leveraging cross-variate information. However, \citet{zeng2023transformers} found that multivariate models underperform simple univariate linear models on many commonly used forecasting benchmarks, as weak cross-variate correlations may not aid forecasting.

Our trapezoidal multi-time series module and multivariate time series input are similar in the way of data concatenation: multiple series are concatenated together in time alignment to form a two-dimensional input. However, it effectively solves the aforementioned weak-correlation problem in multi-time series forecasting. There are two main differences:

\begin{enumerate}
    \item The inherent similarity of same-channel data provides strong cross-series correlation, demonstrating the advantage of multi-series over single-series inputs.
    \item The info length of the series is not equivalent. Unlike traditional multivariate inputs with uniform length, trapezoidal multi-series inputs form a regular matrix while preserving each LTV series' original varying lengths. 
\end{enumerate}

\section{Model Design}

A challenge in LTV forecasting is the series' volatile nature: erratic and frequent fluctuations create non-smoothed temporal patterns, making accurate forecasting particularly difficult. To cope with such challenge, we propose a multi-tower structure called MT-FusionNet, enhancing the robustness of the model to heterogeneous data. MT-FusionNet supports both static and dynamic covariates, utilizing covariate information to improve prediction results. We also design a new training loss function, named the Utilitarian Loss, which corresponds to our trapezoidal multi-time series module, and better addresses the SILO challenge.

\subsection{MT-FusionNet}
\begin{figure*}[t]
\centering
\includegraphics[width=\textwidth]
{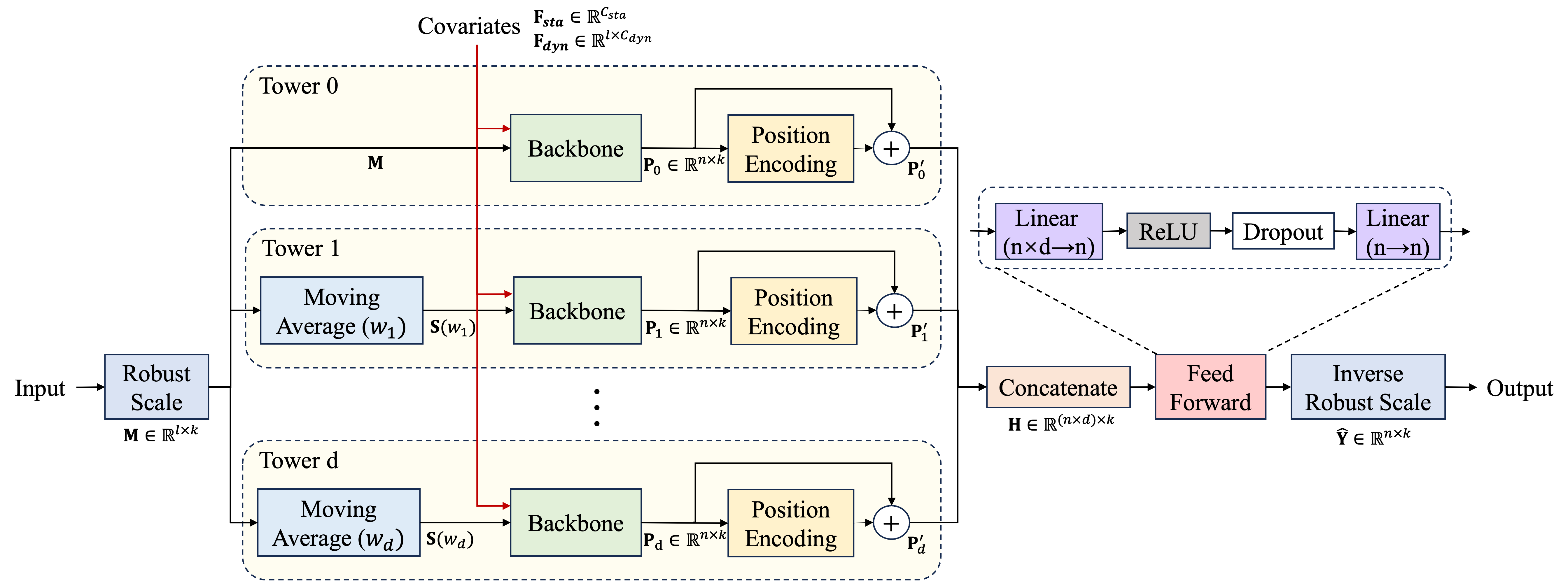} 
\caption{Overview of MT-FusionNet. The input is first normalized by a robust scale. The normalized input is passed through a moving average module of different scales (tower 0 has a moving average scale = 1) and then fed into independent backbones. All backbones can take the same covariates input. After adding position encoding to the output of each backbone, the outputs of all towers are concatenated and passed to a feed-forward network. The final result is obtained after the inverse robust scale.}
\label{fig: mt_fusionnet}
\end{figure*}

The MT-FusionNet mainly consists of three parts: robust scale, multi-tower network, and concatenation. Robust scale solves the distribution shift problem in the series. The multi-tower fusion network introduces smoothing at different scales to provide multi-scale information. The concatenation part fuses the results generated by the multi-tower fusion network and generates the final prediction. The overview of the structure is shown in figure~\ref{fig: mt_fusionnet}. 

\paragraph{Robust Scale}

We follow the symmetric "normalization-denormalization" structure proposed in RevIN \cite{kim2021reversible}, but use a robust instance scale in our model. In contrast to traditional normalization methods (e.g., Z-score normalization) that calculate normalization parameters based on the mean and standard deviation of the data, robust scale uses the median instead of the mean, and interquartile range (IQR, the difference between the 75th and 25th percentiles) instead of the standard deviation. The median is not sensitive to outliers, and IQR only reflects the degree of dispersion of the middle 50\% of the data, so it can more robustly represent the central trend and range of the data. 

For a time series $\mathbf{x} = [x_1, x_2, ..., x_n]$, the scaled series is:
\begin{equation}
\tilde{x}_t = \frac{x_t - \underset{t}{\mathrm{median}}(x_t)}{\mathrm{IQR}} = \frac{x_t - \underset{t}{\mathrm{median}}(x_t)}{Q_3(\mathbf{x}) - Q_1(\mathbf{x})},
\end{equation}
where $Q_3(\mathbf{x})$ denotes the 75th percentile and $Q_1(\mathbf{x})$ denotes the 25th percentile.

\paragraph{Moving Average and Multi Tower}

In order to deal with frequent oscillations in the data, we introduce moving averages of different scales to smooth the normalized data. The moving average improves prediction accuracy by smoothing the data to suppress noise and highlight core patterns. In addition, it reduces the variance of the data, resulting in a more concentrated distribution of the input. 

Time series data often exhibit multi-scale characteristics (e.g., daily fluctuations, weekly cycles, and monthly trends)~\cite{wangtimemixer}. A moving average with a single window size can only capture patterns at one specific temporal granularity. In contrast, a multi-scale moving average utilizing windows of varying sizes can improve forecasting performance by integrating features across different time scales. Smaller windows (e.g., size = 3) are more responsive to short-term variations and retain high-frequency components.
Larger windows (e.g., size = 21) provide stronger smoothing effects and are more suitable for identifying long-term trends. 

Suppose that there are $d$ different smooth scales: $\mathbf{w} = [w_1, w_2, ..., w_d] \in \mathbb{Z}_+^{n}$. To ensure that the shape of the data is consistent before and after smoothing, we use symmetric replicate padding to keep the length. For the input window $\mathbf{M}_c^t$, the smoothed window $\mathbf{S}(w_q)_c^t$ with moving average scale $w_q$ is:
\begin{equation}
\begin{aligned}
\mathbf{S}(w_q)_c^t[i,j] = \frac{1}{w_q} \sum_{r=0}^{w_q-1} \mathbf{M}_c^t[\text{clip}(i + r - \lfloor \tfrac{w_q}{2} \rfloor, 1, l), j],\\ \forall i \in [1,l], j \in [1,k]
\end{aligned}
\end{equation}

\begin{equation}
\text{where} \quad
\text{clip}(t, a, b) = 
\begin{cases} 
a & \text{if } t < a; \\
t & \text{if } a \leq t \leq b; \\
b & \text{if } t > b. 
\end{cases}
\end{equation}

After moving average at different scales, we let the different smoothed data and the original data pass through backbone models, respectively. These models have the same hyperparameter configuration, but their parameters are non-shared and updated independently. When training, these backbones are in parallel positions in the overall model structure, which is called a multi-tower fusion network. 

With $d$ different scales, there are $d+1$ towers in total. The first tower processes the original input $\mathbf{M}_c^t$, while the rest $d$ towers process the different smoothed input $\mathbf{S}(w_q)_c^t$. For simplicity, we omit the start time $t$ and channel $c$ below and use $\mathbf{M}$ and $\mathbf{S}(w_q)$ instead.

\begin{equation}
\mathbf{P}_q = 
\begin{cases} 
\mathtt{Model_0}(\mathbf{M}) & \text{if } q = 0; \\
\mathtt{Model_q}(\mathbf{S}(w_q)) & \text{if } 1 \leq q \leq d.
\end{cases}
\end{equation}
The shape of the output of every tower is the same as the final output $\mathbf{P}_q\in \mathbb{R}^{n \times k}$.

Here, we use the mlp-based model called TSMixer ~\cite{chen2023tsmixerallmlparchitecturetime} as the backbone. However, further experiments show that the proposed multi-tower structure is effective with other mlp-based models. The details will be discussed in the experiment section.

\paragraph{Concatenation}

The outputs of $d+1$ multi-tower structures are concatenated according to the time dimension. To retain the temporal relations of $d+1$ outputs after concatenation, we compute the positional encoding of the outputs and add to the series. The position encoding method follows the Sinusoidal Positional Encoding proposed in Transformer ~\cite{vaswani2017attention}. 
\begin{equation}
\mathbf{P}_q' = \mathbf{P}_q  +\mathtt{Position\_Encoding}(\mathbf{P}_q), \forall q \in [0, d].
\end{equation}
\begin{equation}
\mathbf{H} = \mathtt{Concatenate}([\mathbf{P}_0', \mathbf{P}_1'..., \mathbf{P}_d'], \ \text{dim} = -1) \in \mathbb{R}^{(n \times d) \times k}
\end{equation}

Then, the concatenated result is used to fuse the information of multiple towers through a simple feedforward network (FFN) to reduce the size of the time dimension back to the target time step $n$ of the task.
\begin{equation}
\begin{split}
& \mathbf{H}' = \mathtt{Linear}(n \times d, d)(\mathbf{H}). \\
& \mathbf{H}'' = \mathtt{Dropout}(\mathtt{Activation}(\mathbf{H}')).\\
& \mathbf{\hat{Y}} = \mathtt{Linear}(d,d)(\mathbf{H}'').
\end{split}
\end{equation}

\subsection{Covariates}
The actual LTV curve has prominent peaks or troughs on special dates (e.g., holidays). We introduce auxiliary information to improve the learning of the model for these special points, including static covariates $\mathbf{F}_{sta}$ and dynamic covariates $\mathbf{F}_{dyn}$~\cite{LIM20211748}.

Static covariates $\mathbf{F}_{sta} \in \mathbb{R}^{C_{sta}}$ are covariates that do not change over time and mark the inherent information of the input window. In the user growth scenario, the channel to which each window belongs is the static covariate. The dynamic covariates $\mathbf{F}_{dyn} \in \mathbb{R}^{l \times C_{dyn}}$ change over time and are available ahead of time on the forecast horizon. We introduce two types of dynamic covariates. The first kind is the intrinsic time characteristics, including the features of day/week/month/year. The second is holiday labeling, where we add one-hot labels to holidays based on the observation of the existing LTV curve.

\subsection{Utilitarian Loss}

Based on the trapezoidal multi-time series module, we propose a utilitarian loss for model training. 
In the trapezoidal multi-time series module, what we actually care about is the prediction result $\mathbf{\hat{Y}}_c^t[:,k]$ of the last series $\mathbf{M}_c^t[:,k]$ in the input window.

The traditional MSE loss computes the average error for all series in the window. However, in trapezoidal multi-time series module, the info lengths in the same window are different, and the model is more likely to learn the trend of series with longer info lengths.  This creates an inconsistency problem between the optimization and the actual objective, introducing additional errors.


Intuitively, the utilitarian loss only calculates the MSE loss of the last series, and sets the optimization objective of the model as the actual objective.
\begin{equation}
\mathcal{L}(\mathbf{\hat{Y}}_c^t, \mathbf{Y}_c^t) = \frac{1}{n} \sum_{i=1}^{n} \left( \mathbf{\hat{Y}}_c^t[i,k] - \mathbf{Y}_c^t[i,k]\right)^2
\end{equation}

\section{Experiment}

In this section, we perform experiments to evaluate TTF on real-world industrial data. We implemented the experiment on an internal platform called Merlin in Bytedance. Merlin is a machine learning platform that includes data management, model training, experiment management, service deployment, and automatic iteration, covering the entire process from model development to deployment. 

\subsection{Datasets}

The data set for this experiment comes from Douyin’s user growth business. There are 60 selected user acquisition channels in descending order of the number of acquired users, such as pre-installation on new mobile devices, app store download, etc. We use data from 2021.1.1 to 2023.4.1 as training data and data from 2023.4.1 to 2024.4.1 as test data. As mentioned in covariates, we use two types of feature: the intrinsic time characteristics (day/week/month/year) and one-hot holiday labeling.

\subsection{Experiment Setup}

For each channel on a certain activation date, given $LTV_c^t\text{[0:30]}$, predict $LTV_c^t\text{[30:360]}$, which $LTV_c^t[i]$ represents the $i^{th}$ LTV value of users activated on date $t$ in channel $c$. In order to evaluate the performance of the model on different activation days in each channel, we evaluate the results on 365 activation days. We introduce two metrics to evaluate the results. The first metric is the point-wise mean absolute percentage error($\mathrm{MAPE_{p}}$) of LTV curve , which is used mainly to measure the difference between the predicted curve and the ground truth. We use the number of active users for weighting to eliminate biases of different activation dates and different channels.

\begin{equation}
\mathrm{MAPE_{p}}=\frac{\Sigma_{C,T}(U_c^t \cdot \text{MAPE}(LTV_c^t\text{[30:360]},\hat{LTV_c^t}\text{[30:360]}))}{\Sigma_{C,T} U_c^t},
\end{equation}

where $U_c^t$ is the number of new users activated on date $t$ in channel $c$, $\hat{LTV_c^t}\text{[30:360]}$ is the forecast result and ${LTV_c^t}\text{[30:360]}$ is the ground truth.

The second metric is $\mathrm{MAPE_{a}}$, where $\mathrm{MAPE_{a}}$ denotes the MAPE of LTV\_{N}. We also use the number of active users for weighting.

\begin{equation}
\mathrm{MAPE_{a}}=\frac{\Sigma_{C,T}(U_c^t \cdot \text{MAPE}(\hat{LTV_c^t}\_{N},LTV_c^t\_{N}))}{\Sigma_{C,T} U_c^t},
\end{equation}

where $N=360$, and $U_c^t$ is the number of new users activated on date $t$ in channel $c$, and $\hat{LTV_c^t}\_{N}$ is the forecast result, and ${LTV_c^t}\_{N}$ is the ground truth.

\subsection{Main Results}

We compare our method with the previously deployed online model, which is a statistical framework with expert knowledge and rules. It can achieve reliable prediction performance and has been in production for many years. We also try TSmixer, TiDE and DLinear MLP-based models as the backbone of our structure. 
      
\begin{table}[t]
  \centering
  \caption{Different Backbones. We test the effectiveness of the MT-FusionNet under different MLP-based backbones: TSMixer~\cite{chen2023tsmixerallmlparchitecturetime}, TiDE~\cite{das2024longtermforecastingTiDEtimeseries}, and Dlinear~\cite{zeng2023transformers}. Baseline is a statistical framework with expert knowledge and rules.}
  \begin{tabular}{llcc}
    \toprule
    \textbf{Backbone} & \textbf{MT-FusionNet} & $\mathbf{MAPE_{p}}$ & $\mathbf{MAPE_{a}}$ \\
    \midrule
    \multirow{2}{*}{TSMixer} & w/o & 14.2\% & 9.8\% \\
    & w & \textbf{13.5\%} & \textbf{9.2\%} \\
    \midrule
    \multirow{2}{*}{TiDE} & w/o & 18.2\% & 13.3\% \\
    & w & \textbf{15.6\%} & \textbf{10.8\%} \\
    \midrule
    \multirow{2}{*}{DLinear} & w/o & 29.8\% & 22.6\% \\
    & w & \textbf{24.1\%} & \textbf{20.6\%} \\
    \midrule
    Statistical Model & / & 16.1\% & 11.1\% \\
    \bottomrule
  \end{tabular}
\label{tab:different backbone}
\end{table}

Table \ref{tab:different backbone} presents a comparative evaluation of several backbone models, TSMixer, TiDE, and DLinear, on the LTV\_360 prediction task, with and without MT-FusionNet. Performance is assessed using two metrics: $\mathrm{MAPE_{p}}$ and $\mathrm{MAPE_{a}}$. Across all backbone architectures, the incorporation of the MT-FusionNet consistently yields improved predictive performance. Specifically, the TSMixer with MT-FusionNet achieves the lowest $\mathrm{MAPE_{p}}$ (13.5\%) and $\mathrm{MAPE_{a}}$ (9.2\%) among all evaluated models, outperforming both its origin version ($\mathrm{MAPE_{p}}$: 14.2\%, $\mathrm{MAPE_{a}}$: 9.8\%) and the statistical model ($\mathrm{MAPE_{p}}$: 16.1\%, $\mathrm{MAPE_{a}}$: 11.1\%). Similarly, TiDE and DLinear models benefit from the MT-FusionNet, exhibiting reductions in both $\mathrm{MAPE_{p}}$ and $\mathrm{MAPE_{a}}$ compared to their origin versions (TiDE: $\mathrm{MAPE_{p}}$ reduced from 18.2\% to 15.6\%, $\mathrm{MAPE_{a}}$ from 13.3\% to 10.8\%; DLinear: $\mathrm{MAPE_{p}}$ from 29.8\% to 24.1\%, $\mathrm{MAPE_{a}}$ from 22.6\% to 20.6\%).

These results demonstrate that the MT-FusionNet provides a generalizable improvement across different backbone models for LTV forecasting. Notably, TSMixer with the MT-FusionNet not only surpasses the statistical model, but also outperforms other deep learning backbones.

\begin{figure*}[htbp]
    \centering 
    \begin{subfigure}[b]{0.32\textwidth}
        \centering
        \includegraphics[width=\linewidth]{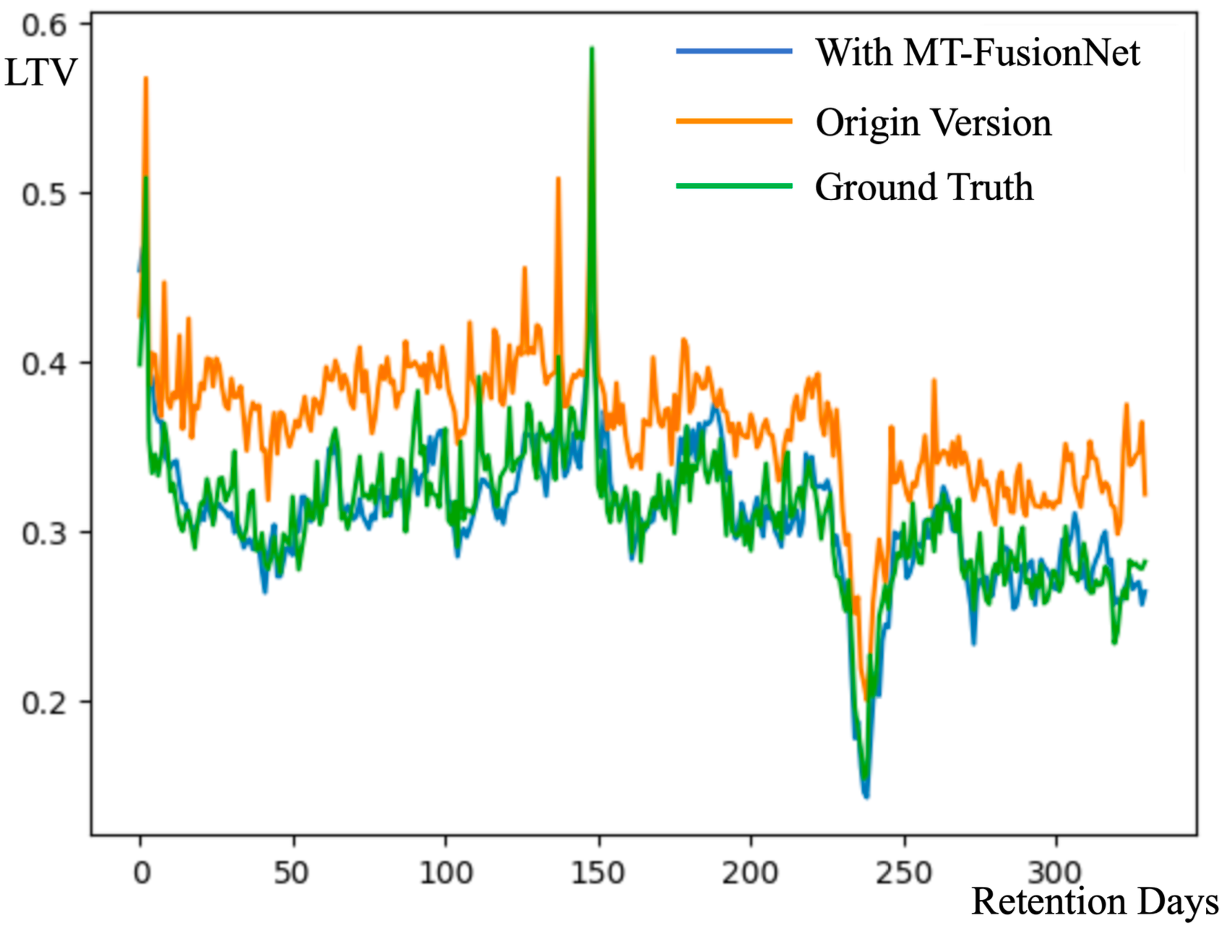}
        \caption{TSMixer}
        \label{fig:sub1}
    \end{subfigure}
    \hfill 
    \begin{subfigure}[b]{0.315\textwidth}
        \centering
        \includegraphics[width=\linewidth]{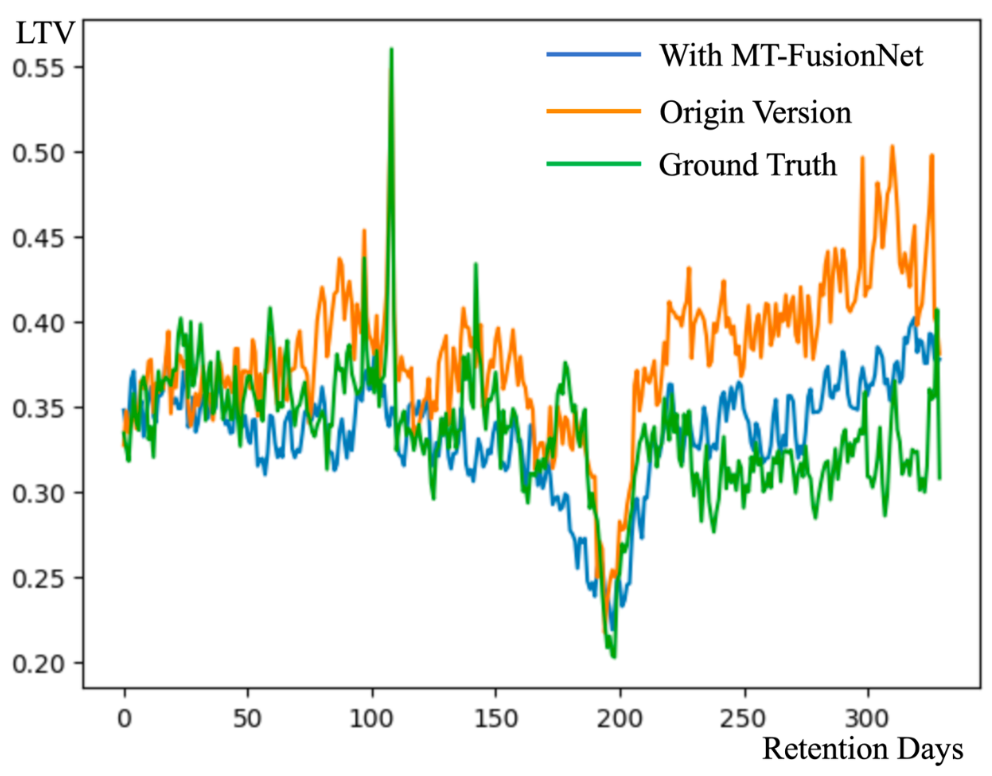}
        \caption{TiDE}
        \label{fig:sub2}
    \end{subfigure}
    \hfill 
    \begin{subfigure}[b]{0.32\textwidth}
        \centering
        \includegraphics[width=\linewidth]{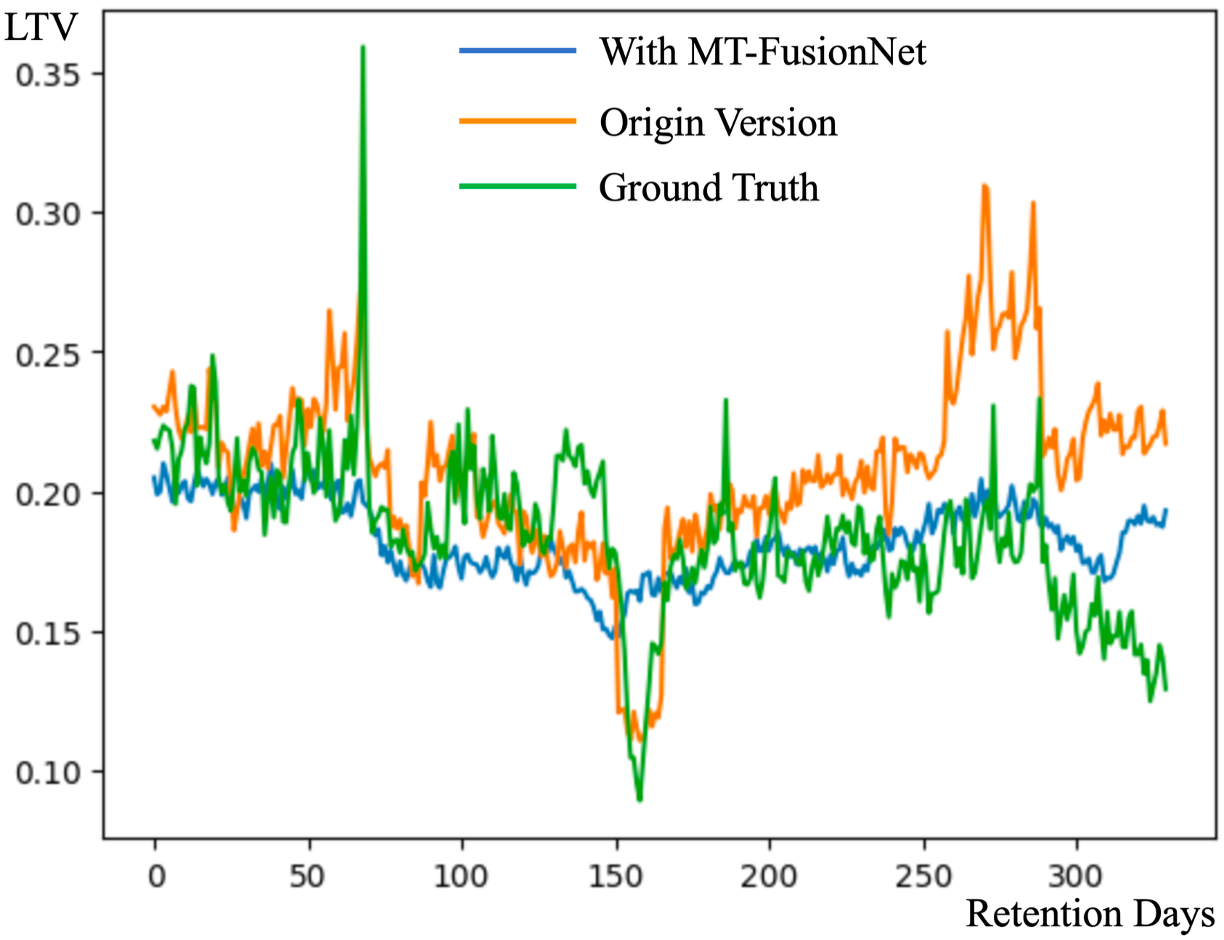}
        \caption{DLinear}
        \label{fig:sub3}
    \end{subfigure}
    \caption{
    Prediction results of with different backbones, with and without the MT-FusionNet. Panels (a)–(c) correspond to TSMixer, TiDE, and DLinear, respectively. Blue: with MT-FusionNet; orange: origin version; green: ground truth. The x-axis denotes retention days; the y-axis denotes LTV. Across the full retention-day horizon, the MT-FusionNet variants align more closely with the ground truth than their origin version, which exhibit a systematic positive bias and a larger phase lag around local extrema. The MT-FusionNet design reduces bias, better tracks short-term fluctuations, and preserves the long-term trend. TSMixer with the MT-FusionNet outperforms the other deep learning backbones.
    }
    \label{fig:total}
\end{figure*}

\subsection{Result Analysis}

Fig.~\ref{fig:total} presents a comparative prediction of several backbone models: TSMixer, TiDE, and DLinear, with MT-FusionNet and its origin versions. These results demonstrate that MT-FusionNet provides a generalizable improvement across different backbone models. TSMixer with MT-FusionNet not only surpasses the statistical model but also outperforms other backbone models.


As Fig. \ref{fig:sub1} shows that TSMixer with MT-FusionNet better follows the pronounced mid-to-late horizon trough (around day 250–270) and its recovery. The amplitude and timing of local peaks are closer to the ground truth, indicating better calibration and reduced lag. In Fig. \ref{fig:sub2}, TiDE with MT-FusionNet yields visibly tighter adherence to the ground truth, particularly mitigating the late-horizon overshoot of the vanilla model. Around the early-horizon spike (60–100 days), the MT-FusionNet captures the rise more faithfully. The largest qualitative gains appear on DLinear (Fig. \ref{fig:sub3}). Without the MT-FusionNet, predictions are rigid and biased upward, failing to capture the gradual downward drift after 200 days.

\subsection{Ablation Study and Parameters Analysis}

In this section, we conduct ablation experiments to evaluate the impact of different modules in our model. The experiments are carried out into four parts: input structure, multi-tower structure, utilitarian loss, and position embedding. We also employ $\mathrm{MAPE_{p}}$ and $\mathrm{MAPE_{a}}$ to evaluate performance. 

\begin{table}[h]
  \centering
  \caption{Performance with different input in TTF framework. Single represents a single series input with length $m=30$. Trapezoidal input has a different number of series in one window: $k = \{180, 360\}$.}
 \begin{tabular}{p{3.75cm} >{\centering\arraybackslash}p{2cm} >{\centering\arraybackslash}p{2cm}} 

    \toprule
    \textbf{Input} & $\mathbf{MAPE_{p}}$ & $\mathbf{MAPE_{a}}$ \\
    \midrule
    Single  & 15.7\% & 12.3\% \\
    Trapezoidal ($k=180$) & \textbf{13.5\%} & \textbf{9.2\%} \\
    Trapezoidal ($k=360$) & 15.6\% & 11.9\% \\
    \bottomrule
  \end{tabular}
  \label{tab:input structure}
\end{table}



\begin{table}[h]
  \centering
  \caption{Performance of different moving average scale settings $\mathbf{w}$.}
  \begin{tabular}{lcc}
    \toprule
    \textbf{Moving Average Scale 
    }  & $\mathbf{MAPE_{p}}$ & $\mathbf{MAPE_{a}}$ \\
    \midrule
    $\mathbf{w}$ = [1] & 14.2\% & 9.8\% \\
    $\mathbf{w}$ = [1,3,7] & 13.9\% & 9.5\% \\
    $\mathbf{w}$ = [1,3,5,7] & 14.1\% & 9.9\% \\
    $\mathbf{w}$ = [1,3,7,14] & \textbf{13.5\%} & \textbf{9.2\%} \\
    $\mathbf{w}$ = [1,7,14,21] & 14.3\% & 9.9\% \\
    $\mathbf{w}$ = [1,3,14,21] & 13.6\% & 9.3\% \\
    $\mathbf{w}$ = [1,3,5,7,14] & 14.2\% & 9.8\% \\
    $\mathbf{w}$ = [1,3,5,7,9,11,13,15] & 13.8\% & 9.3\% \\
    \bottomrule
  \end{tabular}
  \label{tab:Multi-Tower Structure}
\end{table}
As shown in Table \ref{tab:input structure}. The results demonstrate that trapezoidal input has a number of 180 series in one window and outperforms both the single-dimension strategy and the 360 series in one window. This suggests that while multi-dimensional data organization is beneficial, increasing the number of activation dates beyond an optimal point (180 in this case) can lead to performance degradation, possibly due to overfitting or increased model complexity without proportional information gain. 

In Table \ref{tab:Multi-Tower Structure}, we can see that the tower configuration [1,3,7,14] demonstrated superior performance with the lowest  $\mathrm{MAPE_{p}}$ (13.5\%) and $\mathrm{MAPE_{a}}$ (9.2\%). This configuration appears to capture an optimal balance of feature representations at different scales. Compared to the [1] configuration, it achieves a 4.9\% relative improvement in $\mathrm{MAPE_{p}}$ and a 6.1\% improvement in $\mathrm{MAPE_{a}}$. The results indicate that simply increasing the number of towers does not necessarily lead to improved performance and combinations of small and large towers can effectively capture both different patterns in the data. 


\begin{table}[h]
  \centering
  \caption{Performance comparison between the Utilitarian Loss and MSE Loss.}
  \begin{tabular}{p{3cm} >{\centering\arraybackslash}p{2cm} >{\centering\arraybackslash}p{3cm}} 

    \toprule
    \textbf{Loss type} & $\mathbf{MAPE_{p}}$ & $\mathbf{MAPE_{a}}$ \\
    \midrule
    Utilitarian Loss & \textbf{13.5\%} & \textbf{9.2\%} \\
    MSE Loss & 13.8\% & 9.4\% \\
    \bottomrule
  \end{tabular}
  \label{tab:Utilitarian Loss}
\end{table}
We can see from Table \ref{tab:Utilitarian Loss} that the utilitarian loss function works better than the MSE loss function on both performance metrics, with a 0.3\% improvement in $\mathrm{MAPE_{p}}$ (13.8\% to 13.5\%) and a 0.2\% improvement in $\mathrm{MAPE_{a}}$ (9.4\% to 9.2\%). The utilitarian loss function can reduce the noise of different activation dates prediction of the target sequence and make a better prediction. 


\begin{table}[t]
  \centering
  \caption{Performance comparison between MT-FusionNet with and without layer Position Embedding in each tower.}
  \begin{tabular}{p{3cm} >{\centering\arraybackslash}p{2cm} >{\centering\arraybackslash}p{3cm}} 

    \toprule
    \textbf{Position Encoding} & $\mathbf{MAPE_{p}}$ & $\mathbf{MAPE_{a}}$ \\
    \midrule
    with & \textbf{13.5\%} & \textbf{9.1\%} \\
    w/o & \textbf{13.5\%} & 9.2\% \\
    \bottomrule
  \end{tabular}
  \label{tab:Position Embedding}
\end{table}
As for the experiment of position embedding. In table \ref{tab:Position Embedding} the result indicates that embedding has an impact on model performance. Although $\mathrm{MAPE_{p}}$ remains unchanged at 13.5\% for both configurations, there is a slight improvement in $\mathrm{MAPE_{a}}$ (9.1\%  vs 9.2\%, a 0.1\% improvement) when position embedding is used. This suggests that position embedding provides a benefit in capturing sequential patterns or positional information in the data, particularly for long-term predictions.

\section{Industrial deployment}

\begin{figure}[t]
\centering
\includegraphics[width=\linewidth]{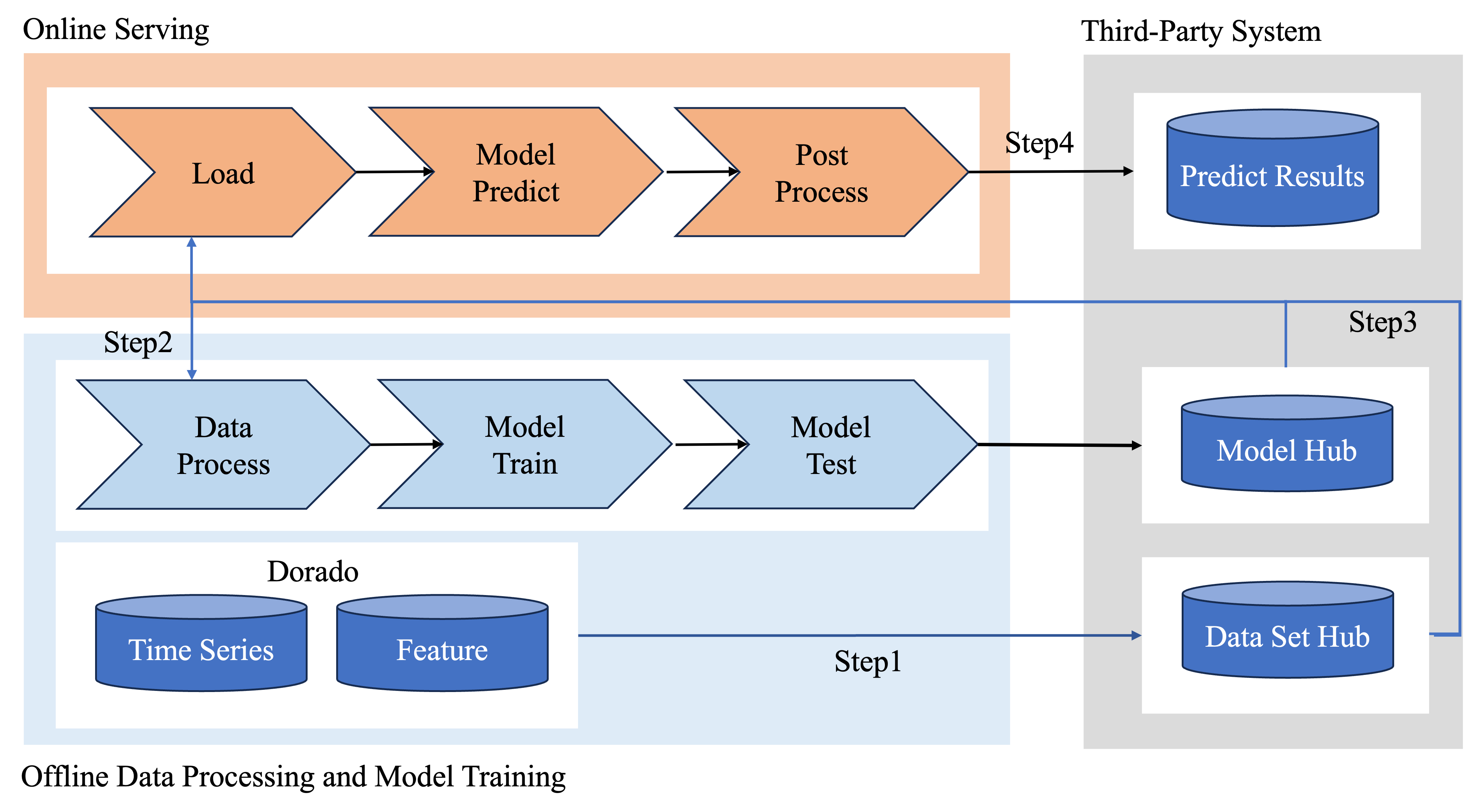} 
\caption{ Overview of system 
deployed in a production environment. The system is organized by three subsystems: (i) Offline data processing and model training, responsible for data processing, model training, and model evaluation; (ii) Online serving handles prediction requests by loading versioned datasets and model artifacts, running inference, and performing post‑processing; (iii) Third‑Party system integrates external repositories that manage datasets, model artifacts, and prediction results.}
\label{fig: online structure}
\end{figure}

Our proposed framework has been deployed in a production environment system since March of this year. As illustrated in Figure~\ref{fig: online structure}, the system is organized into three subsystems: (i) Offline data processing and model training, responsible for data processing, model training, and model evaluation; (ii) Online serving handles prediction requests by loading versioned datasets and model artifacts, running inference, and performing post‑processing; (iii) Third‑Party system integrates external repositories that manage datasets, model artifacts, and prediction results. The end‑to‑end workflow is summarized in four steps.

\begin{itemize}
\item Data preprocessing and data set publication. Raw time series and features are prepared on Dorado, a one‑stop big‑data platform at ByteDance. The Data Process module performs cleaning, temporal alignment and resampling, normalization, feature encoding, and data set splitting. The curated datasets, together with their schemas, feature dictionaries, and transformation specifications, are versioned and published in the Data Set Hub. 

\item Data loading, training and evaluation. Training jobs load the specified data set versions from the Data Set Hub. Models are trained in the Model Train component and evaluated in the Model Test component using cross‑validation and predefined metrics. When acceptance criteria are met, the trained model is serialized and registered, along with its metadata, in the Model Hub.

\item Model loading and online inference. In online serving, the loading stage retrieves the approved model artifact from the Model Hub and the required inference inputs from the Data Set Hub. The Model Predict stage executes forward inference, followed by post‑process operations.

\item Persistence of results. The final predictions are written into the Predict Results repository in the third‑party system. Each record includes the prediction value with its associated identifiers, as well as the model and data set versions to support auditing and rollback.
\end{itemize}

Compared to the previously deployed online model,after three months of online deployment, $\mathrm{MAPE_{p}}$ decreased by 4.3\%, and $\mathrm{MAPE_{a}}$ decreased by 3.2\%. 

The system provides channel-level LTV predictions to support marketing strategists and budget planners. To ensure operational stability, a robust MLOps workflow is implemented. Models are automatically retrained weekly, with additional retraining cycles triggered by a monitoring system after detecting a significant performance drift (e.g., a \>2\% increase in MAPE over a 7-day window). An automated alerting system notifies the on-call team of such degradation or data pipeline failures. For governance, auditability is maintained by logging each prediction with version identifiers for the model, data, and code. Furthermore, a rollback mechanism enables immediate reversion to a previously validated model in response to critical failures, ensuring business continuity.

\section{Conclusion}

In this paper, we study the early-stage channel-level LTV forecasting problem in user growth scenarios. It is fundamentally different from conventional time series forecasting due to the unaligned multi-time series, the SILO constraint, and the volatile, non-stationary characteristics of real LTV data. We proposed the Trapezoidal Temporal Fusion (TTF) framework, incorportating a trapezoidal multi-time series module to mitigate data unalignment and the SILO challenge, and MT-FusionNet to enhance prediction accuracy. TTF framework has been successfully implemented in the online production system for Douyin, resulting in substantial performance improvements, with $\mathrm{MAPE_{p}}$ and $\mathrm{MAPE_{a}}$ reduced by 4.3\% and 3.2\% respectively. These results demonstrate the effectiveness and practical value of TTF framework for large-scale, real-world LTV forecasting tasks, and suggest its potential applicability to other domains facing similar unaligned multi-time series forecasting challenges.

\clearpage
\bibliography{aaai2026}

\end{document}